\documentclass[letterpaper]{article} 
\usepackage{aaai24}  
\usepackage{times}  
\usepackage{helvet}  
\usepackage{courier}  
\usepackage[hyphens]{url}  
\usepackage{graphicx} 
\usepackage{natbib}  
\usepackage{caption} 
\usepackage{algorithm}
\usepackage{algorithmic}

\usepackage{amssymb}

\usepackage{newfloat}
\usepackage{listings}
\DeclareCaptionStyle{ruled}{labelfont=normalfont,labelsep=colon,strut=off} 
\floatstyle{ruled}
\newfloat{listing}{tb}{lst}{}
\floatname{listing}{Listing}
\title{Fine Structure-Aware Sampling: A New Sampling Training Scheme\\ for Pixel-Aligned Implicit Models in Single-View Human Reconstruction}

\author {
    Kennard Yanting Chan\textsuperscript{\rm 1,\rm 2},
    Fayao Liu\textsuperscript{\rm 2},
    Guosheng Lin\textsuperscript{\rm 1},
    Chuan Sheng Foo\textsuperscript{\rm 2,\rm 3},
    Weisi Lin\textsuperscript{\rm 1}
}

\affiliations {
    \textsuperscript{\rm 1}Nanyang Technological University\\
    \textsuperscript{\rm 2}Institute for Infocomm Research, A*STAR, Singapore\\
    \textsuperscript{\rm 3}Centre for Frontier AI Research, A*STAR, Singapore\\
    kenn0042@e.ntu.edu.sg, liu\_fayao@i2r.a\mbox{-}star.edu.sg, gslin@ntu.edu.sg, foo\_chuan\_sheng@i2r.a\mbox{-}star.edu.sg, wslin@ntu.edu.sg
}

\usepackage{bibentry}

\usepackage{float}
\usepackage{subcaption}


\begin{document}

\maketitle

\begin{abstract}
Pixel-aligned implicit models, such as PIFu, PIFuHD, and ICON, are used for single-view clothed human reconstruction. These models need to be trained using a sampling training scheme. Existing sampling training schemes either fail to capture thin surfaces (e.g. ears, fingers) or cause noisy artefacts in reconstructed meshes. To address these problems, we introduce Fine Structured-Aware Sampling (FSS), a new sampling training scheme to train pixel-aligned implicit models for single-view human reconstruction. FSS resolves the aforementioned problems by proactively adapting to the thickness and complexity of surfaces. In addition, unlike existing sampling training schemes, FSS shows how normals of sample points can be capitalized in the training process to improve results.
Lastly, to further improve the training process, FSS proposes a mesh thickness loss signal for pixel-aligned implicit models. It becomes computationally feasible to introduce this loss once a slight reworking of the pixel-aligned implicit function framework is carried out. Our results show that our methods significantly outperform SOTA methods qualitatively and quantitatively. Our code is publicly available at https://github.com/kcyt/FSS.
\end{abstract}

\section{Introduction}

3D reconstruction of human bodies is an area that has garnered interest due to its potential applications in fields such as virtual reality, 3D printing, and game production. Although it is already possible to accurately reconstruct a human body using high-end, multi-view capturing systems \cite{collet2015high,lombardi2018deep}, such systems are unavailable to typical consumers. This has led to research efforts to develop deep learning models for 3D human reconstruction using sparse inputs like a single RGB image \cite{alldieck2019learning,natsume2019siclope,saito2020pifuhd}.

An influential class of deep learning methods for single-image clothed human reconstruction is pixel-aligned implicit models \cite{saito2019pifu,saito2020pifuhd, chan2022integratedpifu}. These methods learn an implicit function that represents the surface of a human body. From the learned implicit function, a mesh of a human body can be extracted using the Marching Cubes algorithm \cite{lorensen1987marching}.

\begin{figure}[t]
\centering
\includegraphics[width=0.5\textwidth]{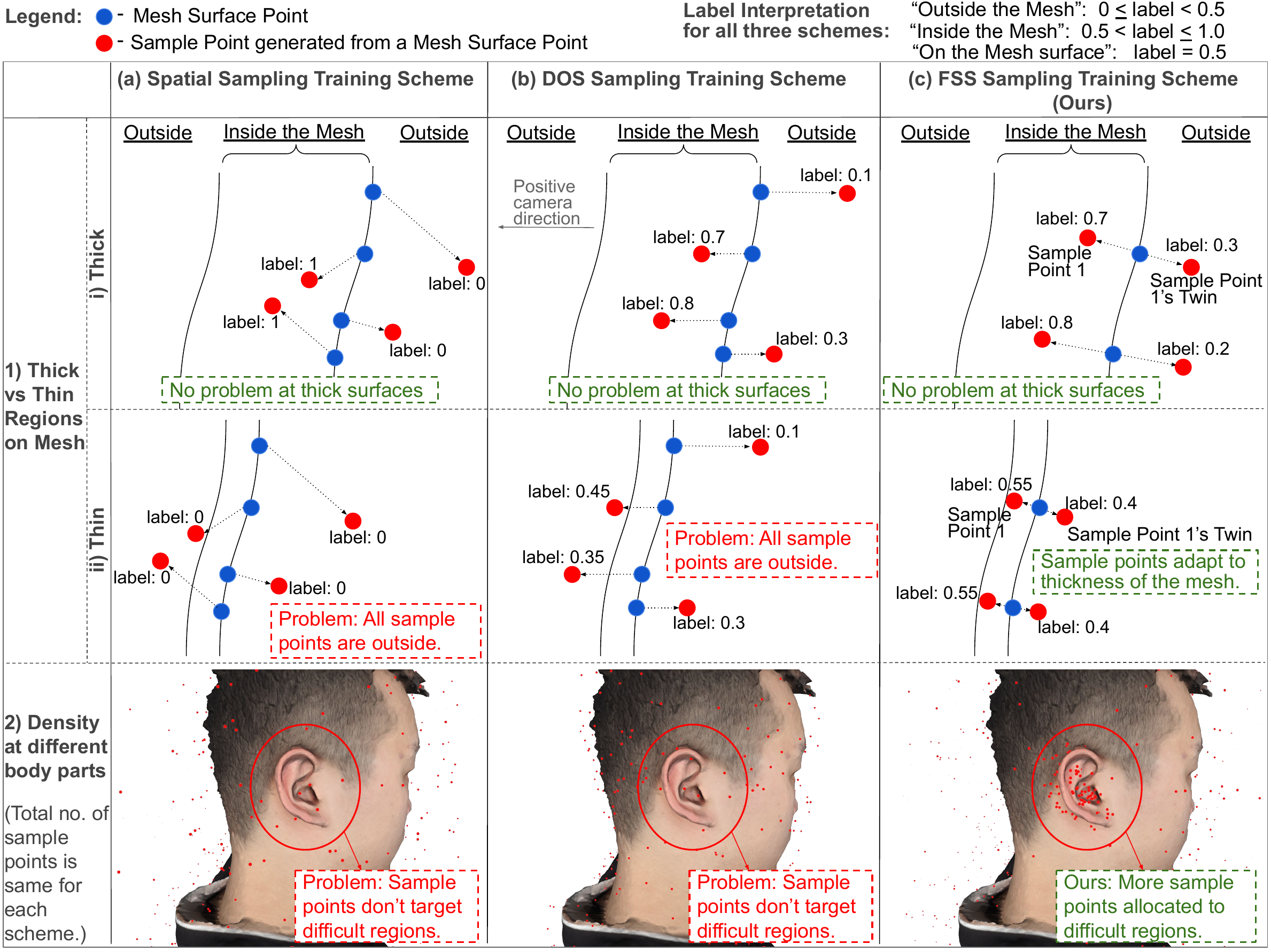}
\vspace{-7mm}
\caption{Unlike existing schemes, FSS can: 1. Adapts to thickness of mesh. 2. Prioritize regions that are challenging.}
\label{fig:SmplxGuidedSampling}
\vspace{-7mm}
\end{figure}

\begin{figure*}[t]
\centering
\includegraphics[width=0.90\textwidth]{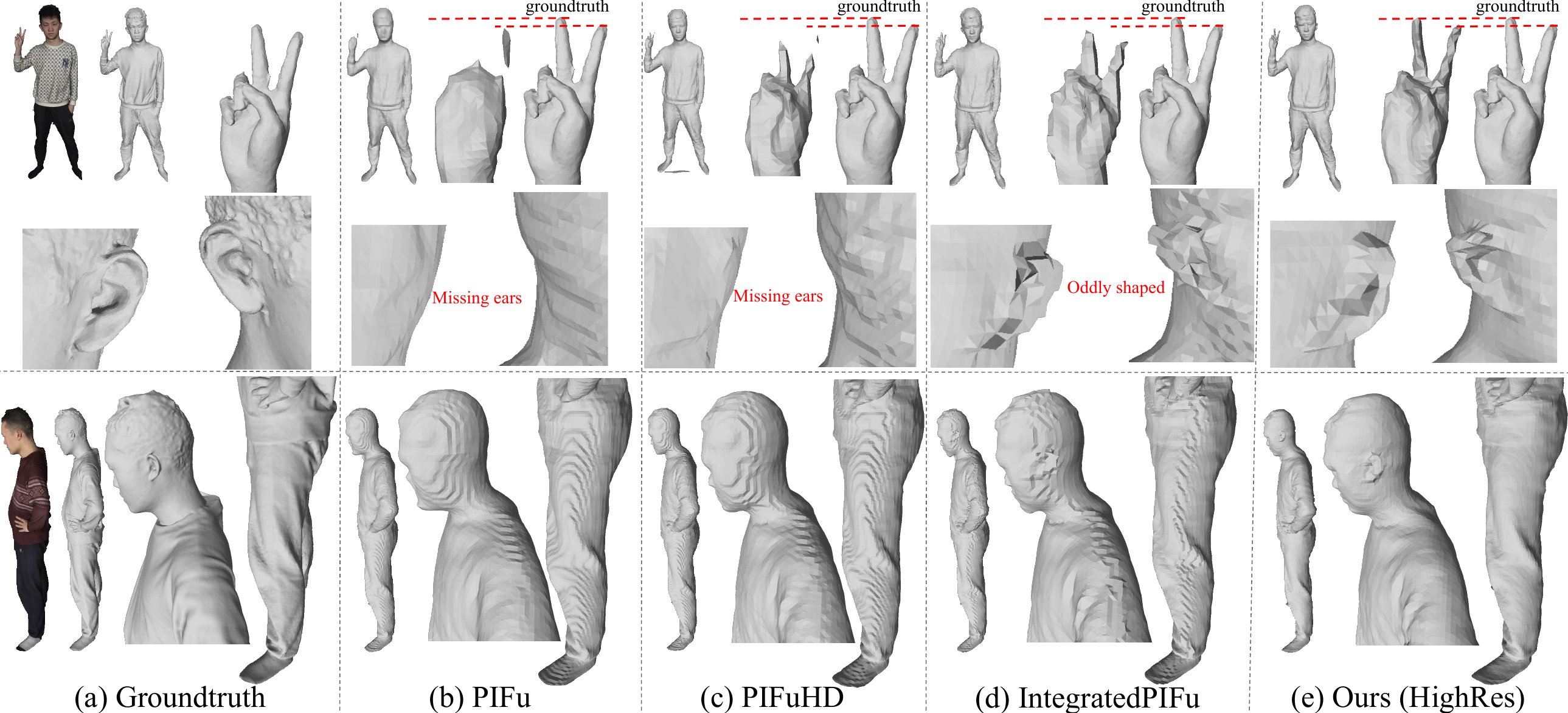}
\vspace{-2.85mm}
\caption{Unlike SOTA methods, our method captures thin body features (e.g. fingers, ears) w/o causing noisy, wavy artefacts.}
\label{fig:firstPageSota}
\vspace{-6.45mm}
\end{figure*}

To learn the implicit function, all pixel-aligned implicit models have to be trained using a sampling training scheme. A sampling training scheme provides supervision signals to a pixel-aligned implicit model by generating sample points and their corresponding labels. How the sample points are generated and what labels are computed will drastically affect the models' results. The earliest sampling training scheme, called spatial sampling scheme, was introduced by the first pixel-aligned implicit model - PIFu \cite{saito2019pifu}. Since then, subsequent pixel-aligned implicit models, such as PIFuHD \cite{saito2020pifuhd}, PaMIR \cite{zheng2021pamir}, ICON \cite{xiu2022icon}, S-PIFu \cite{chans}, and more, simply continue to use the same scheme.

IntegratedPIFu \cite{chan2022integratedpifu} is the first work to contribute a new sampling training scheme (i.e. Depth-Oriented Sampling or DOS) that differs materially from spatial sampling. Unlike spatial sampling, DOS is able to train models to reconstruct thin but important body features like ears and fingers. But DOS suffers from a lack of robustness. Specifically, DOS only helps camera-facing mesh surfaces, and causes wavy, noisy artefacts on non-camera-facing surfaces. 

To overcome these issues, we propose \textbf{Fine Structure-aware Sampling (FSS)}, a new sampling training scheme that teaches pixel-aligned implicit models to reconstruct thin body features that are not only artefact-free but also structurally accurate. FSS achieves this by proactively adapting to the thickness and complexity of surfaces. This is shown in Fig. \ref{fig:SmplxGuidedSampling}, which also compares FSS with the existing schemes.

In addition, we offer two extensions to our FSS scheme. The first of which is the exploitation of the Normals of Sample Points \textbf{(NSP)}. Unlike existing sampling training schemes, NSP shows how normals of sample points can be capitalized on during training to further improve reconstruction results. We will elaborate more later.

The second extension to FSS is Mesh Thickness Loss \textbf{(MTL)} signal. Although MTL is not related to the sample points generated by FSS, we consider MTL to be an extension of FSS as MTL will affect the supervision signals sent by FSS to a pixel-aligned implicit model. It is well-publicized that a pixel-aligned implicit model tends to produce reconstructed meshes that have body parts with implausible thickness \cite{hong2021stereopifu,peng2021neural,chan2022integratedpifu}. We argue that this problem can be solved if a pixel-aligned implicit model is able to directly learn the thickness of different body parts. Thus, we propose a mesh thickness loss  function, which is our MTL, to do that. Introducing this loss is non-trivial as predicted meshes (and its thickness) are not available during train time. We will introduce our solution later.

In short, we contribute: 1. FSS, a robust scheme that captures thin, important body features. 2. NSP, an extension to FSS that incorporates normals of sample points into training process. 3. MTL, another extension that incorporates mesh thickness loss to supervision signals produced by FSS.

\section{Related Work}

\subsection{Single-view Human Reconstruction}
Methods that reconstruct a human body mesh from a single image can be broadly classified into two classes: Parametric methods and non-parametric methods. 

Parametric methods \cite{kanazawa2018end,pang2022benchmarking,pang2024towards} recover human body shapes by predicting parameters belonging to a chosen human parametric model (e.g. SMPL-X \cite{SMPL-X:2019}). However, these parametric methods can only produce cloth-less and hairless human body meshes. While methods like Multi-garment Net \cite{bhatnagar2019multi} and Bcnet \cite{jiang2020bcnet} try to predict clothes on top of a human parametric model, these methods are unable to produce accurate clothed meshes.

On the other hand, non-parametric methods do not use a human parametric model. As aforementioned, a subclass of non-parametric methods that has attracted significant attention from the research community is the pixel-aligned implicit models. PIFu \cite{saito2019pifu} is the first of such models, and it is able to reconstruct highly accurate clothed human meshes from a single image.

After PIFu, other pixel-aligned implicit models have been proposed. These include PIFuHD \cite{saito2020pifuhd}, StereoPIFu \cite{hong2021stereopifu}, S-PIFu \cite{chans}, IntegratedPIFu \cite{chan2022integratedpifu}, and more.

\begin{figure*}[t]
\centering
\includegraphics[width=0.85\textwidth]{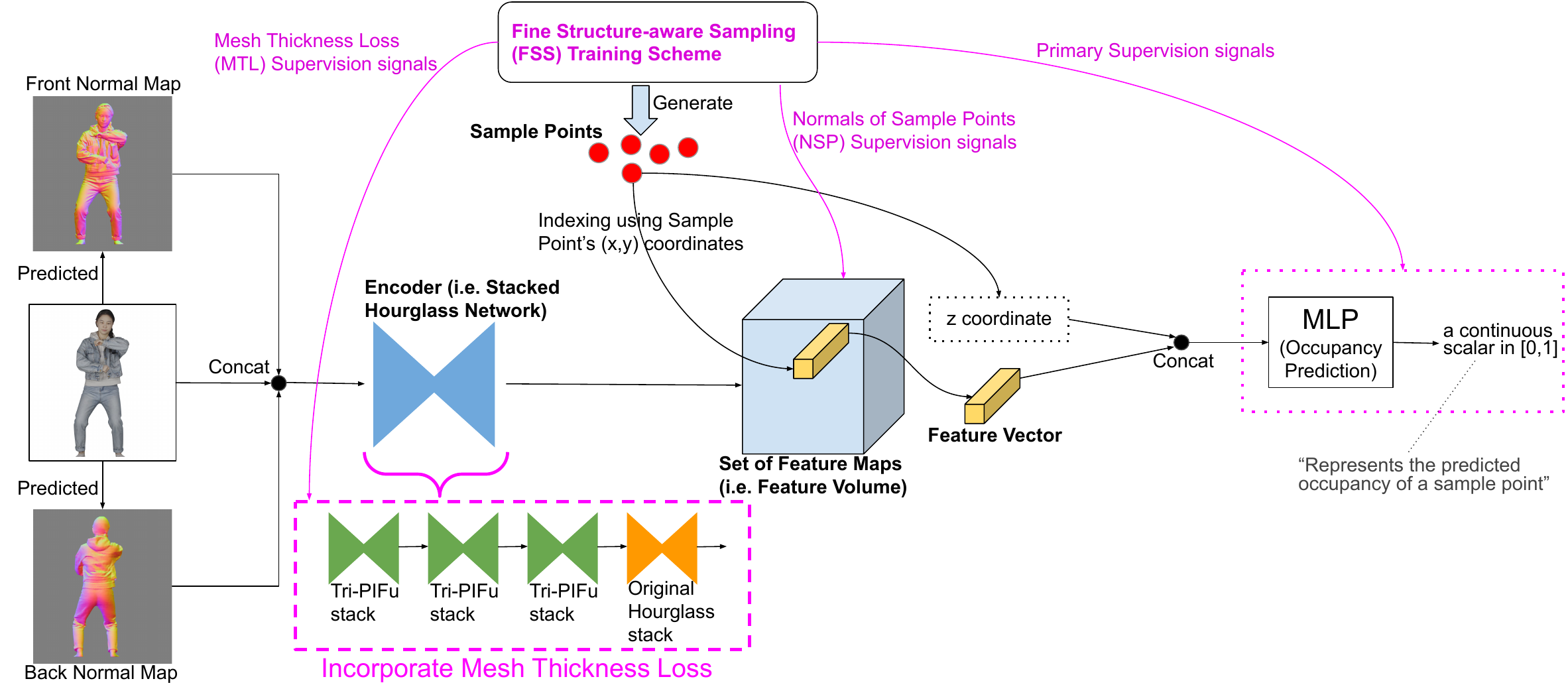}
\vspace{-3mm}
\caption{Overview of our FSS sampling training scheme (with NSP and MTL) in a single architecture.}
\label{fig:methodDiagram}
\vspace{-7mm}
\end{figure*}

\subsection{PIFu and IntegratedPIFu}
\label{PIFuAndIntegratedPIFu_section}

The original PIFu \cite{saito2019pifu} uses an encoder-decoder architecture. An illustration of the PIFu architecture is given in Fig. \ref{fig:methodDiagram} (Not including the predicted normal maps and the magenta arrows, fonts, and boxes). The encoder is a 2D CNN (usually a stacked hourglass network \cite{newell2016stacked} consisting of 4 stacks). From a RGB image, the encoder produces a set of feature maps (i.e. feature volume). During training and testing, 3D sample points are sampled within the 3D camera space (of the RGB image). The (x,y) coordinates of the sample points are then used to index the feature volume to obtain feature vectors, with each feature vector corresponding to a sample point. The z coordinate of the sample point, together with the feature vector obtained by the sample point, would then be passed to the decoder, which is a multi-layer perceptron (MLP). The MLP will predict a value from 0 to 1 where value$>0.5$ means the sample point is `inside' a groundtruth human body mesh, value$<0.5$ means the sample point is `outside', and value=0.5 means the sample point is exactly on a human body surface. During testing, the 3D camera space of a RGB image is evenly sampled to get a 3D grid of evenly spaced sample points. By predicting these sample points' values, we get a 3D grid of values. By applying the Marching Cube algorithm \cite{lorensen1987marching} on this grid of values, we obtain a reconstructed 3D human mesh.

During training, PIFu uses what is known as the spatial sampling scheme to generate sample points and assign labels to these sample points. The spatial sampling scheme mainly generates sample points by displacing mesh surface points of a groundtruth human mesh with normally-distributed noise. These sample points can be displaced in any direction and in any magnitude. An illustration of these sample points is shown in Fig. \ref{fig:SmplxGuidedSampling}a. Sample points in this scheme are given binary labels (either 0 or 1).  

In IntegratedPIFu \cite{chan2022integratedpifu}, the authors identified problems with the spatial sampling scheme and proposed the Depth-Oriented Sampling (DOS) training scheme. They showed that the binary labels (as opposed to continuous labels) and unconstrained displacement of mesh surface points in spatial sampling scheme caused thin but obvious body features like the ears and fingers to be missing in the reconstructed meshes. Their proposed DOS scheme introduced continuous labels and constrained displacement of surface points. Continuous labels provide more information and clearer supervision to a pixel-aligned implicit model during training. Constrained displacement of mesh surface points (i.e. surface points are only displaced in the camera-direction) narrows down the meaning of each label, and this makes learning simpler for a pixel-aligned implicit model. These two advantages allow a pixel-aligned implicit model to reconstruct human meshes that retain thin body features like ears and fingers, unlike spatial sampling. However, DOS has serious flaws that will be explained in the next section.

\section{Method}

Most pixel-aligned implicit models, including PIFuHD, StereoPIFu, S-PIFu, and IntegratedPIFu, use PIFu as a major building block(s) in their own architectures. Thus, we test our FSS scheme in an architecture and pipeline (see Fig. \ref{fig:methodDiagram}) that are identical to those of a PIFu, except for the use of predicted normal maps, FSS scheme, and Tri-PIFu stacks.  

First, like in PIFuHD and IntegratedPIFu, we use a pix2pixHD network \cite{wang2018high} to predict front and back normal maps from an input RGB image. The normal maps and RGB image are concatenated and fed into an encoder, which produces a feature volume. Next, sample points will index the feature volume to retrieve feature vectors that are fed into a MLP, which will then predict the occupancy of these sample points.

Fig. \ref{fig:methodDiagram} also shows an overview of our three contributions. First, FSS trains a pixel-aligned implicit model to reconstruct thin body features by giving clearer and more meaningful `Primary Supervision signals', which are used to guide occupancy prediction. Second, an extension of FSS is to use Normals of Sample Points (NSP) as additional supervision signals. NSP signals guide and mold the representations in the feature volume. Third, another extension of FSS is to introduce Mesh Thickness Loss (MTL) supervision signals, which will train the encoder to account for mesh thickness. We will now elaborate on each of the three.

\subsection{Fine Structure-aware Sampling Scheme (FSS)}
\label{ProximityguidedSamplingScheme_section}

As explained in Section \ref{PIFuAndIntegratedPIFu_section}, DOS is a sampling training scheme that was recently proposed in IntegratedPIFu \cite{chan2022integratedpifu} to succeed the spatial sampling scheme.

To teach pixel-aligned implicit models to reconstruct thin body features (e.g. ears, fingers), DOS constrains sample points' labels to only the camera direction (see Fig. \ref{fig:SmplxGuidedSampling}b), making it easier for models to interpret and learn from the labels. For example, if a sample point has a label of 0.7 in DOS, it means that a short distance away (in positive or negative camera-direction) from this sample point, we will find a mesh surface. Conversely, if labels are not constrained to camera-direction only, then a label of 0.7 does not really pinpoint where the mesh surface is (as any direction is likely). 

But because the sample points' labels are determined by the shortest distance between a sample point and a mesh surface in the camera direction, rather than the shortest distance in any direction, a DOS-trained pixel-aligned implicit model is trained to only identify surfaces that are front-facing and back-facing. Lateral surfaces become difficult for a DOS-trained model to identify. Thus, DOS suffered from problems such as high-frequency, wavy artefacts on the side-facing surfaces of its reconstructed human meshes. Moreover, badly reconstructed side-facing surfaces also mean thin body features (e.g. ears, fingers) tend to get incorrect shapes.

To overcome the drawbacks of DOS, we propose FSS. FSS solves the above issues with its \textbf{5 key features}.
 
\begin{figure}[t]
\centering
\includegraphics[width=0.5\textwidth]{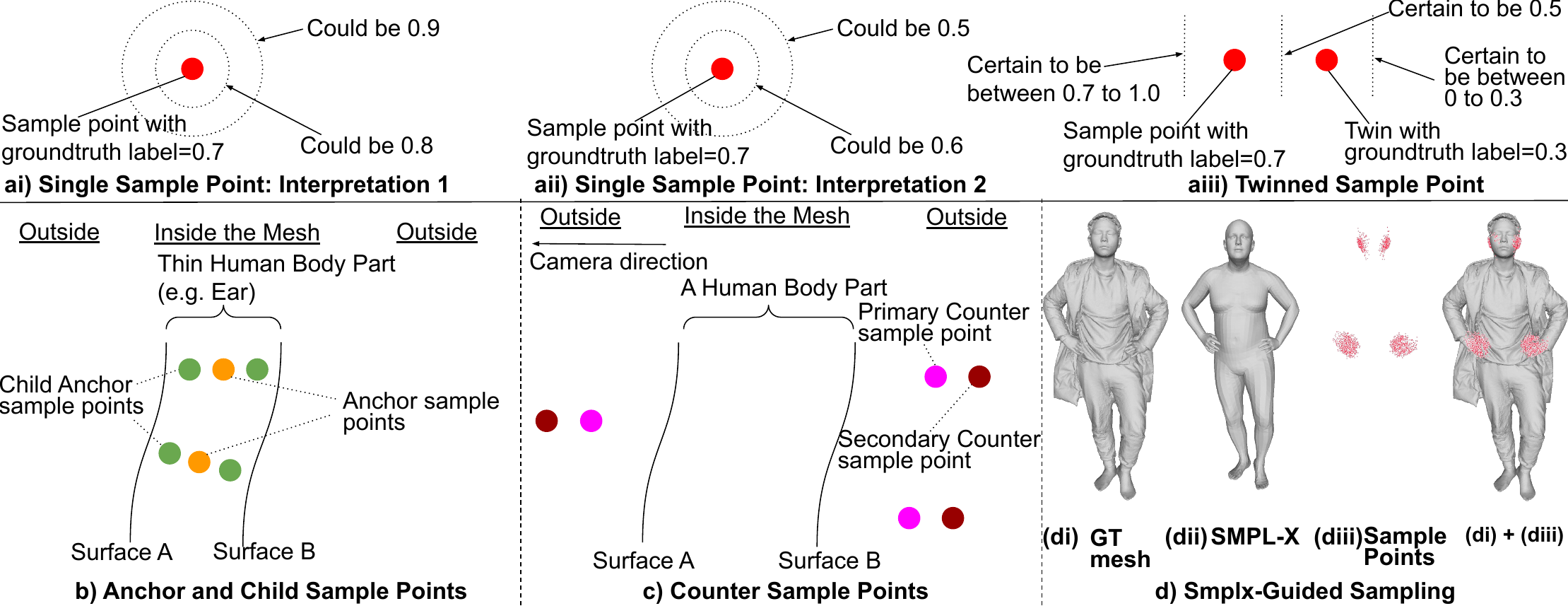}
\vspace{-7mm}
\caption{Four of the five key features in FSS. (a) Twinned Sample Points (b) Anchor Sample Points (c) Counter Sample Points (d) Smplx-guided Sampling.}
\label{fig:SmplxGuidedSampling_AnchorPts_and_CounterPts_and_SmplxGuiding}
\vspace{-7mm}
\end{figure}

\vspace{-1mm}
\paragraph{1. Twinned Sample Points}

In order to have labels that are determined by the shortest distance in any direction (unlike DOS) and yet still makes it easy to pinpoint the actual location of mesh surfaces, FSS proposes twinned sample points.

In both spatial sampling and DOS schemes, the sample points are generated independently from one another. But in FSS, our sample points will each have a corresponding twin (see Fig. \ref{fig:SmplxGuidedSampling}c ). As explained earlier, a sample point with a label determined by shortest distance from any direction cannot unambiguously pinpoint the location of a nearby mesh surface. This is illustrated in Fig. \ref{fig:SmplxGuidedSampling_AnchorPts_and_CounterPts_and_SmplxGuiding}ai \& Fig. \ref{fig:SmplxGuidedSampling_AnchorPts_and_CounterPts_and_SmplxGuiding}aii, which show the same sample point having two different interpretations.

However, this can be overcome if we have a pair of sample points (twins) that are equidistant away from the nearby mesh surface (See Fig. \ref{fig:SmplxGuidedSampling_AnchorPts_and_CounterPts_and_SmplxGuiding}aiii). In other words, the midpoint of the pair of twins automatically pinpoints the location of a mesh surface. Being able to precisely pinpoint the mesh surface is pivotal for learning to reconstruct thin body features like ears or fingers. Moreover, as seen in last row of Fig. \ref{fig:SmplxGuidedSampling}a-b, only a few sample points are near thin surfaces (i.e. ears) each time. It is thus vital to make full use of every point.

\vspace{-1mm}
\paragraph{2. Proximity-adaptive Displacement}

Other than twinned sample points, FSS also introduces the concept of a proximity-adaptive displacement. Proximity-adaptive displacement is motivated by the observation that pixel-aligned implicit models trained with spatial sampling scheme rarely encounter any sample point that fall within thin surfaces like ears or fingers.	This is because, as shown in Fig. \ref{fig:SmplxGuidedSampling}, a sample point is generated by displacing a mesh surface point with some random noise. Thus, compared to mesh surface points on thicker surfaces, mesh surface points on thin surfaces are much more likely to be displaced into a region outside of the mesh (See the first and second rows in Fig. \ref{fig:SmplxGuidedSampling}a and \ref{fig:SmplxGuidedSampling}b). With the majority of sample points that are near thin surfaces labelled as `outside’ (i.e. label$<$0.5), this naturally encourages the pixel-aligned implicit models to predict `outside’ for any sample points near thin surfaces, resulting in missing ears, fingers, and other thin body parts in reconstructed meshes. Proximity-adaptive displacement overcomes the issue by adjusting how much to displace the mesh surface point based on thickness of the surfaces (see top two rows in Fig. \ref{fig:SmplxGuidedSampling}c).

\vspace{-1mm}
\paragraph{3. Anchor Sample Points}

In addition, Fig. \ref{fig:SmplxGuidedSampling}c (second row) shows another problem with capturing thin body features when we use continuous labels. In almost all cases, the maximum label of a sample point inside a thin body feature will never be close to 1.0 due to how thin the body feature is. In our experiments, we find that the maximum label tends to be around 0.60. Thus, with thin body features, the range of sample points' labels is distorted from [0, 1] to [0,  $\sim$0.60].

There is thus a bias towards predicting values less than 0.5 (i.e. `outside’ ) for any sample point close to thin surfaces. To solve this, we can naively increase the number of sample points that are inside thin surfaces. However, that would introduce a large number of additional sample points. Thus, in order to correct the bias efficiently, we propose the idea of anchor sample points (illustrated in Fig. \ref{fig:SmplxGuidedSampling_AnchorPts_and_CounterPts_and_SmplxGuiding}b). Anchor sample points are sample points that are at the deepest location inside a thin body part. These points will have the highest label (e.g. 0.60) that is possible in the thin body part. From an anchor sample point, we generate a number of \textbf{child anchor sample points}. Child anchor sample points are sample points that are in between an anchor sample point and the nearest mesh surface (See Fig. \ref{fig:SmplxGuidedSampling_AnchorPts_and_CounterPts_and_SmplxGuiding}b). In addition to correcting the bias, anchor sample points and child anchor sample points are also important for indicating to the pixel-implicit aligned model where the max label value would be reached, and that the model should start predicting a label value that is lower than the max label value for any sample point located in between an anchor sample point and a child anchor sample point (i.e. clearer supervision signals for the model).

\vspace{-0.75mm}
\paragraph{4. Counter Sample Points}

FSS also includes sample points that deter floating artefacts. Floating artefacts often appears either in front or behind a reconstructed mesh, where `front' or `behind' is determined by the camera-direction. Thus, we propose counter sample points. Counter sample points are points that are either in front or behind the mesh. They are always outside of the mesh and are used to discourage a pixel-aligned implicit model from predicting floating artefacts in regions that are actually empty. We further enhance this with the concept of a ``twinned" counter points. We pair a counter point (i.e. a primary counter point) with a secondary counter point (see Fig. \ref{fig:SmplxGuidedSampling_AnchorPts_and_CounterPts_and_SmplxGuiding}c). The secondary counter point is located further away (in camera direction) from the mesh surface compared to the primary counter point, and thus the secondary counter point will have a lower label than primary counter point. Twinned counter points indicate the direction of where the labels should be decreasing in, thereby providing clearer supervision signals. 

\vspace{-1.0mm}
\paragraph{5. Smplx-guided Sampling}

Finally, the human body is a complex and convoluted structure. To pixel-aligned implicit models, some body parts will be easier to reconstruct than others. Existing sampling training schemes (i.e. spatial sampling and DOS) do not discriminate between these different body parts and do not assign less sample points to body parts that are easier to reconstruct. Examples of such body parts include the neck, lower leg, and chest, which are made of mostly flat surfaces and are easy for models to reconstruct. 

Thus, FSS introduces Smplx-guided sampling, which allows us to select which human body parts we want to focus on (and assign more sample points there). In our context, we have FSS to focus on the important thin human body features (e.g. ears and fingers) that we are interested in reconstructing. Smplx-guided sampling requires the use of a groundtruth SMPL-X mesh \cite{SMPL-X:2019} (only during training). Using the SMPL-X mesh, we can identify the location of ears and fingers of the groundtruth clothed human mesh (See Fig. \ref{fig:SmplxGuidedSampling_AnchorPts_and_CounterPts_and_SmplxGuiding}d). We then generate a higher concentration of sample points from those locations so as to help a pixel-aligned implicit model focus on these body parts. As illustrated in the third row of Fig. \ref{fig:SmplxGuidedSampling}, Smplx-guided sampling reduces the density of sample points on easy-to-reconstruct parts like the human neck and increases the density of sample points on hard-to-reconstruct parts like the human ears.

\subsection{Exploiting Normals of Sample Points (NSP)}

\begin{figure}[t]
\centering
\includegraphics[width=0.40\textwidth]{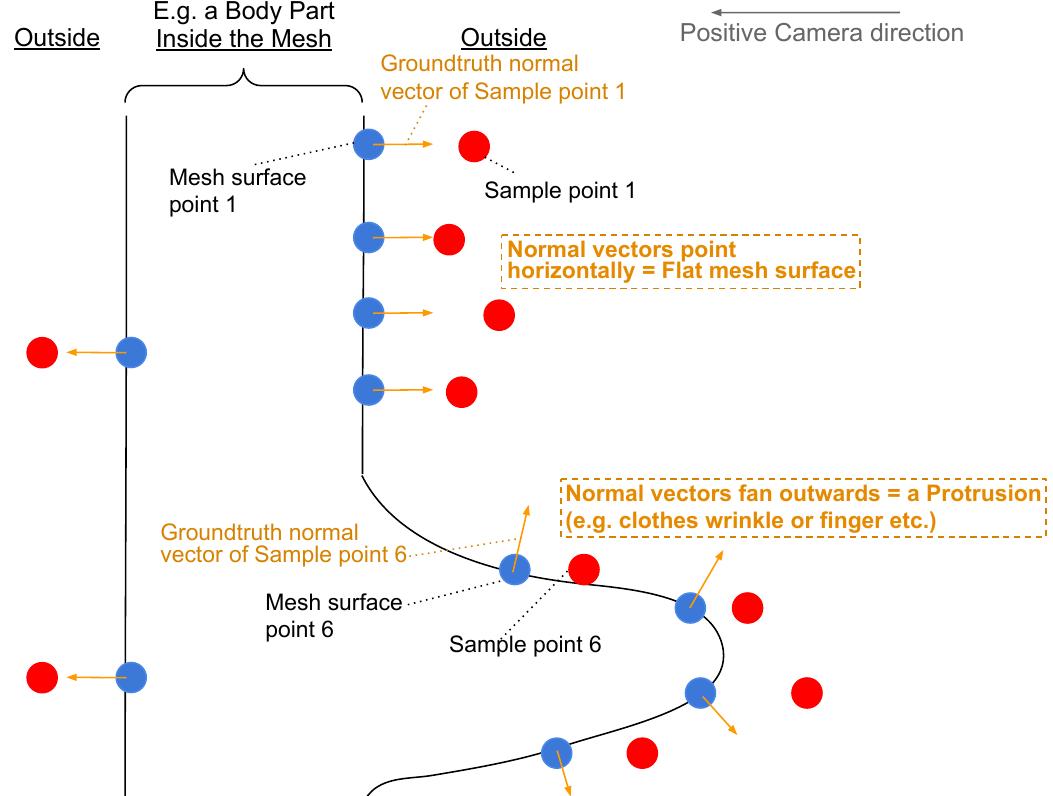}
\vspace{-3.5mm}
\caption{Usefulness of sample points' normals}
\label{fig:ProblemPredNormal}
\vspace{-3.5mm}
\end{figure}

\begin{figure}[t]
\centering
\includegraphics[width=0.5\textwidth]{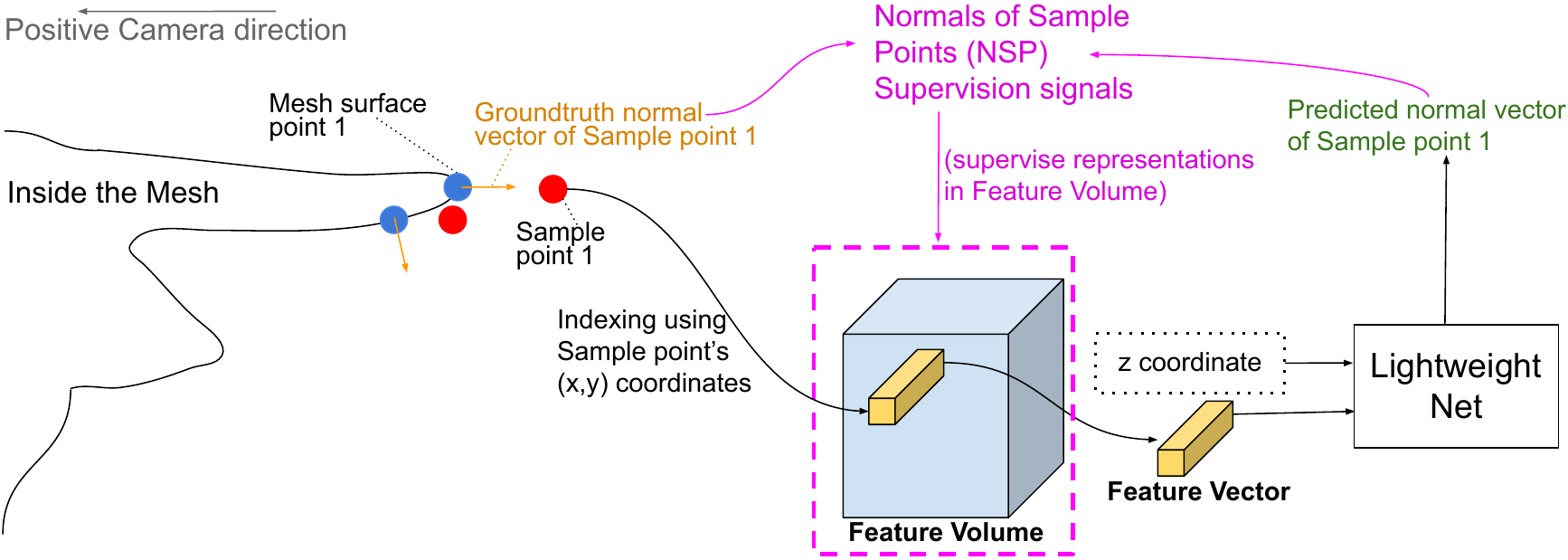}
\vspace{-7.5mm}
\caption{Illustration of NSP, an extension to FSS.}
\label{fig:PredNormal}
\vspace{-8mm}
\end{figure}

The sample points' normals, which can be computed, are regrettably not used in existing sampling training schemes. The closest related work that tried is PHORHUM \cite{alldieck2022photorealistic}. PHORHUM slightly modifies the spatial sampling scheme such that normals of sample points that lie \textbf{exactly on the mesh surface} are used to regularize the training process. In contrast, our proposed idea will use the normals of \textbf{all} our sample points.

In reality, the normals of sample points are very useful for inferring the underlying structure of a human body mesh (see Fig. \ref{fig:ProblemPredNormal}). Hence, as an extension to our FSS scheme, we exploit the normals of sample points to further enhance the training process of pixel-aligned implicit models.

We define the normal of a sample point as the normal of the mesh surface point that is nearest to that sample point in the camera direction (Refer to Fig. \ref{fig:ProblemPredNormal}). 

We use the normals of sample points to supervise and enhance the representations in the Feature Volume in Fig. \ref{fig:methodDiagram}. To do so, we use the sample points to index the Feature Volume (refer to Fig. \ref{fig:PredNormal}) and obtain a set of feature vectors. The feature vector indexed by a sample point, along with the z coordinate of this sample point, will be fed into a small neural network that will attempt to predict the normal vector of this sample point. The mean squared error between the predicted and groundtruth normal vectors is computed, and the derived supervision signals are used to train the representations in the Feature Volume. The small neural network only acts as a dummy as it is only used during training.

A point to highlight is that there is a difference between the predicted normal maps that are used as inputs to a pixel-aligned implicit model and our proposed use of normals of sample points (NSP). While the former directly injects normal information into the model, the latter regularizes and improves the representations in the Feature Volume such that ridges, peaks, and troughs on a clothed body are registered and retained in these representations. The impact of NSP is supported by our ablation studies that will be shown later.

\subsection{Mesh Thickness Loss (MTL) Supervision}

Works, such as \cite{hong2021stereopifu,peng2021neural,chan2022integratedpifu}, observed that pixel-aligned implicit models tend to produce reconstructed meshes with implausible thickness. An intuitive solution to this problem is to teach the pixel-aligned implicit model to directly learn the thickness of different body parts (at different orientations). Thus, FSS proposes a mesh thickness loss (MTL) signal to do that.

It is, however, not trivial to introduce this loss. This is because predicted meshes (and their thicknesses) are not available during train time. During training, for each image, a pixel-aligned implicit model will process and predict the labels of (typically) only 8000 sample points. During testing, to predict a human mesh, the same model would need to process and predict for (typically) $256^3$=16,777,216 sample points. Thus, the time cost alone makes it infeasible to compute the mesh thickness of predicted meshes at train time. Moreover, since the marching cube algorithm used to produce the predicted mesh is not differentiable, it is unclear how we could backpropagate the error between predicted meshes' thickness and groundtruth meshes' thickness.

\begin{figure}[t]
\centering
\includegraphics[width=0.46\textwidth]{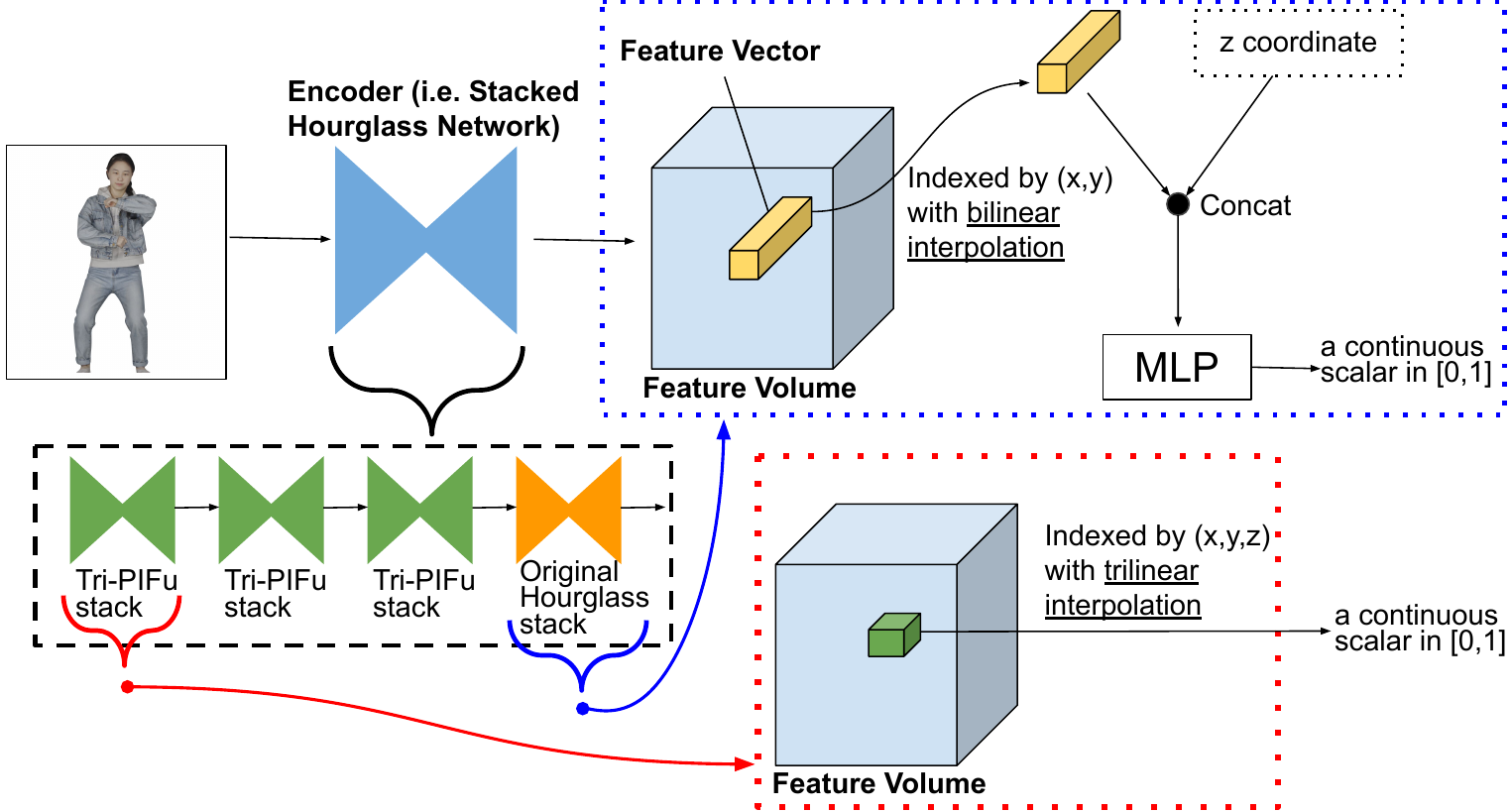}
\vspace{-3.75mm}
\caption{Illustration of our Tri-PIFu architecture}
\label{fig:SDF_Filter}
\vspace{-8mm}
\end{figure}

To solve the issue, we first introduce an architectural modification of the PIFu framework. In Fig. \ref{fig:methodDiagram}, we see that the Encoder is a Stacked Hourglass Network consisting of 4 stacks. In PIFu, unlike what is shown in Fig. \ref{fig:methodDiagram}, 4 Hourglass stacks (in orange) are used. Tri-PIFu stack (in green) is a new type of stack that we proposed to replace 3 of the 4 Hourglass stacks. Each of the 4 stacks, regardless of their types, will produce a separate Feature Volume (see Fig. \ref{fig:SDF_Filter}).

As shown by the blue dotted box in Fig. \ref{fig:SDF_Filter}, an Hourglass stack would produce a Feature Volume that is indexed via bilinear interpolation to retrieve feature vectors. Feature vectors are concatenated with z coordinates and fed into a MLP that predicts occupancy values that range from 0 to 1.

In contrast, as shown by the red dotted box in Fig. \ref{fig:SDF_Filter}, a Tri-PIFu stack does not use the MLP, and we do not concatenate the feature vector with the z coordinate. Instead, we use the sample point's (x,y,z) coordinates to index the Feature Volume via \textbf{trilinear interpolation} to directly retrieve an occupancy value that ranges from 0 to 1.

The Feature Volume produced by a Tri-PIFu stack is interpreted as a 3D space with shape of (D, H, W), where D, H, W represents its depth, height, and width respectively. Tri-PIFu's aim is to model an implicit function of a human mesh surface \textbf{inside} this 3D space. To do so, we applied sigmoid activation function on the Tri-PIFu stack's outputs, thereby ensuring every value or element in the Feature Volume falls within [0,1]. This ensures that any sample point that index the Feature Volume via trilinear interpolation will always obtain a value within [0,1]. Once trained, the stack will model an implicit function (of a predicted human mesh surface) inside its Feature Volume. Aside: The predicted surface is at the 0.5 (not 0) level-set of the implicit function.

With the implicit function in the Feature Volume, it is easy to get a metric of mesh thickness at any (x,y) position. We simply sum up the Feature Volume in the z (or D) dimension to obtain a 2D plane of shape (H, W). Each value or element in this 2D plane would be a consistent approximation of the mesh thickness at that (x,y) position. We refer to this 2D plane as the mesh thickness plane.

We can easily compute the groundtruth mesh thickness plane using the groundtruth mesh. The mean squared error between the groundtruth and predicted mesh thickness planes is then computed during training. This loss signal is our MTL signal. MTL signals are an extension to our FSS scheme as the signals provide additional supervision to the training process. MTL will be ablated later. More on MTL (e.g. why we still use 1 Hourglass stack) in Supp. Mat.

\section{Experiments}

\subsection{Datasets}
In our experiments, we use the THuman2.0 dataset \cite{tao2021function4d} as the training set for both our models and other competing models. THuman2.0 dataset contains 526 high-quality scans (or meshes) of ethnic Chinese human subjects. We use a 80-20 train-test split of these meshes. For each training mesh, we first render a RGB image of the mesh's front view using a weak-perspective camera. We then render 10 other images by evenly fanning out (i.e. changing the yaw) from the first RGB image, in both clockwise and counter-clockwise directions. This set-up is similar to the ones used in IntegratedPIFu \cite{chan2022integratedpifu} and S-PIFu \cite{chans}, which are benchmarks that we aim to compete against in our experiments.

In addition, we made use of the BUFF dataset \cite{Zhang_2017_CVPR} to evaluate all the models. None of the models are trained using the BUFF dataset. Like what the authors did in IntegratedPIFu, we conducted systematic sampling (based on sequence number) on the BUFF dataset, giving us 101 human meshes to be used for evaluating the models. By using systematic sampling, we avoided getting meshes that have both the same human subject and the same pose.

\subsection{Comparison with State-of-the-art}

\begin{figure*}[t]
\centering
\includegraphics[width=0.85\textwidth]{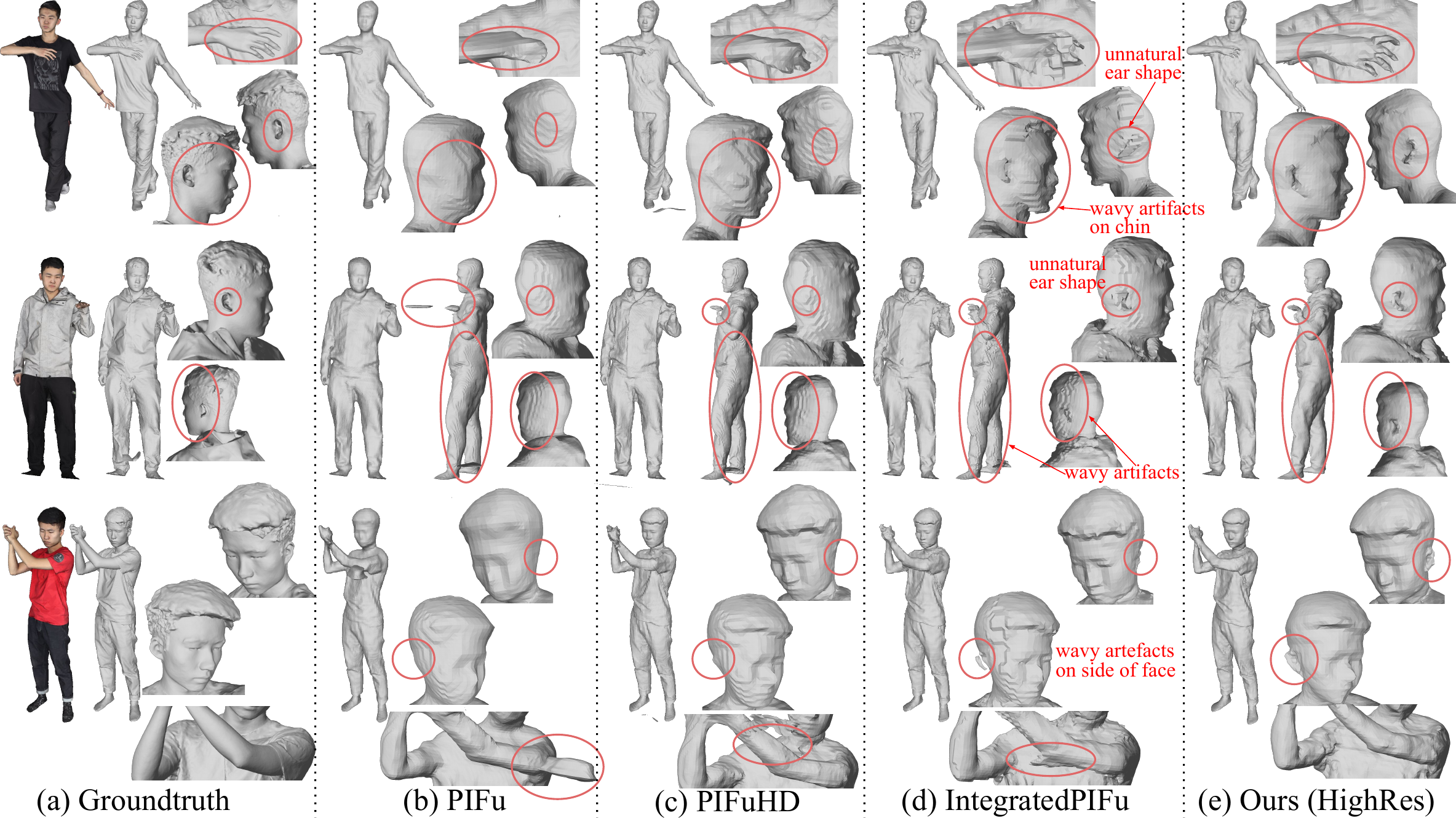}
\vspace{-2.5mm}
\caption{Evaluation with SOTA models.}
\label{fig:secondPageSota}
\vspace{-3.0mm}
\end{figure*}

We trained two models. The first uses the architecture in Fig. \ref{fig:methodDiagram}. The second is same as the first but uses the HRI component proposed by IntegratedPIFu to incorporate high-res images. We compare our models against existing models on single-view clothed human reconstruction. The existing models include PIFu \cite{saito2019pifu}, PIFuHD \cite{saito2020pifuhd}, and IntegratedPIFu \cite{chan2022integratedpifu}. To be fair to all models, predicted normal maps are used in all models, including PIFu. Following \cite{saito2019pifu,saito2020pifuhd, chan2022integratedpifu}, we use Chamfer distance (CD), Point-to-Surface (P2S), and Normal reprojection error (Normal) as the metrics in our quantitative evaluation. In addition, to compare our methods against methods that use SMPL-X meshes as priors, we added S-PIFu \cite{chans} in our quantitative evaluation. Due to space constraint, S-PIFu is compared qualitatively with ours in our Supp. Mat. We also compared against ICON \cite{xiu2022icon} in our Supp. Mat.

\vspace{-1.25mm}
\paragraph{Qualitative Evaluation}
We evaluate the models qualitatively in Fig. \ref{fig:firstPageSota} and Fig. \ref{fig:secondPageSota}. The figures show our method reconstructs fine, thin features like fingers, hands, and ears correctly. Unlike existing models, our method does not produce high-frequency wavy artefacts or unnatural protrusions (i.e. implausible mesh thickness) on reconstructed meshes. 

We also provided a qualitative evaluation based on real Internet images sourced from Shutterstock in Fig. \ref{fig:InternetPhotos}.

\vspace{-1.5mm}
\paragraph{Quantitative Evaluation}

\setlength{\tabcolsep}{4pt}
\begin{table}
\begin{center}
\caption{Our models against SOTA methods. (`H-Res' indicates if a 1024x1024 RGB image is required and used. By default, a 512x512 image is used. HRI=High-Resolution Integrator proposed by IntegratedPIFu, FSS=Fine Structure-aware Sampling, NSP=Trained with Normals of Sample Points, MTL=Trained with Mesh Thickness Loss)}
\vspace{-4mm}
\label{table:sotaQuantitative}
\resizebox{1.0\columnwidth}{!} {
\begin{tabular}{ll|lll|lll}
\hline 
 & & \multicolumn{3}{c|}{THuman2.0 Test Set} & \multicolumn{3}{c}{BUFF} \\
Methods & H-Res & CD (10\textsuperscript{-4}) & P2S (10\textsuperscript{-4}) & Normal (10\textsuperscript{-2}) & CD (10\textsuperscript{3}) & P2S (10\textsuperscript{3}) & Normal (10\textsuperscript{-2}) \\
\hline
PIFu & $\times$ & 5.314 & 4.933 & 8.149 & 2.089 & 1.977 & 6.243 \\ 
PIFu + DOS & $\times$ & 5.943 & 5.794 & 8.224 & 2.214 & 2.077 & 6.393 \\ 
S-PIFu & $\times$ & 5.000 & 4.728 & 8.079 & 2.140 & 2.011 & 6.178 \\ 
ICON (w predicted SMPL-X) & $\times$ & 6.862 & 7.808 & 12.96 & 2.757 & 3.137 & 8.569 \\
PIFuHD & $\checkmark$ & 5.267 & 4.688 & 7.667 & 2.177 & 2.072 & 5.967 \\
IntegratedPIFu (HRI + DOS) & $\checkmark$ & 5.172 & 4.276 & 7.620 & 2.061 & 1.778 & 5.935 \\ 
\hline
FSS w/o NSP w/o MTL (Ours) & $\times$ & 5.004 & 3.965 & 8.052 & 2.001 & 1.737 & 6.056 \\
FSS w/o MTL (Ours) & $\times$ & 4.931 & 3.916 & 7.780 & 1.947 & 1.652 & 5.838 \\
FSS w/o NSP (Ours) & $\times$ & 4.923 & 3.969 & 7.995 & 1.957 & 1.678 & 5.941 \\
FSS (Ours) & $\times$ & {\bf 4.833} & {\bf 3.854} & 7.800 & {\bf 1.943} & {\bf 1.576} & 5.812 \\
HRI + FSS (Ours) & $\checkmark$ & 4.896 & 3.905 & {\bf 7.615} & 1.945 & 1.611 & {\bf 5.715} \\
\hline
\end{tabular} 
}
\end{center}
\vspace{-6.5mm}
\end{table}
\setlength{\tabcolsep}{1.4pt}

We also evaluate our models quantitatively in Tab. \ref{table:sotaQuantitative}. From the table, we can see that our low-resolution model, `FSS (Ours)', is able to outperform the existing models (first five rows) in all except one column. The failure to outperform PIFuHD and IntegratedPIFu in the `Normal' metric for the THuman2.0 dataset can be attributed to PIFuHD's and IntegratedPIFu's use of a higher resolution input RGB image. Indeed, with our high-resolution model, `HRI + FSS (Ours)', we are able to significantly outperform the existing models in all metrics for both datasets.

\subsection{Ablation Studies}
\paragraph{Evaluation of FSS scheme (w/o NSP w/o MTL)}

\begin{figure}[t]
\centering
\includegraphics[width=0.5\textwidth]{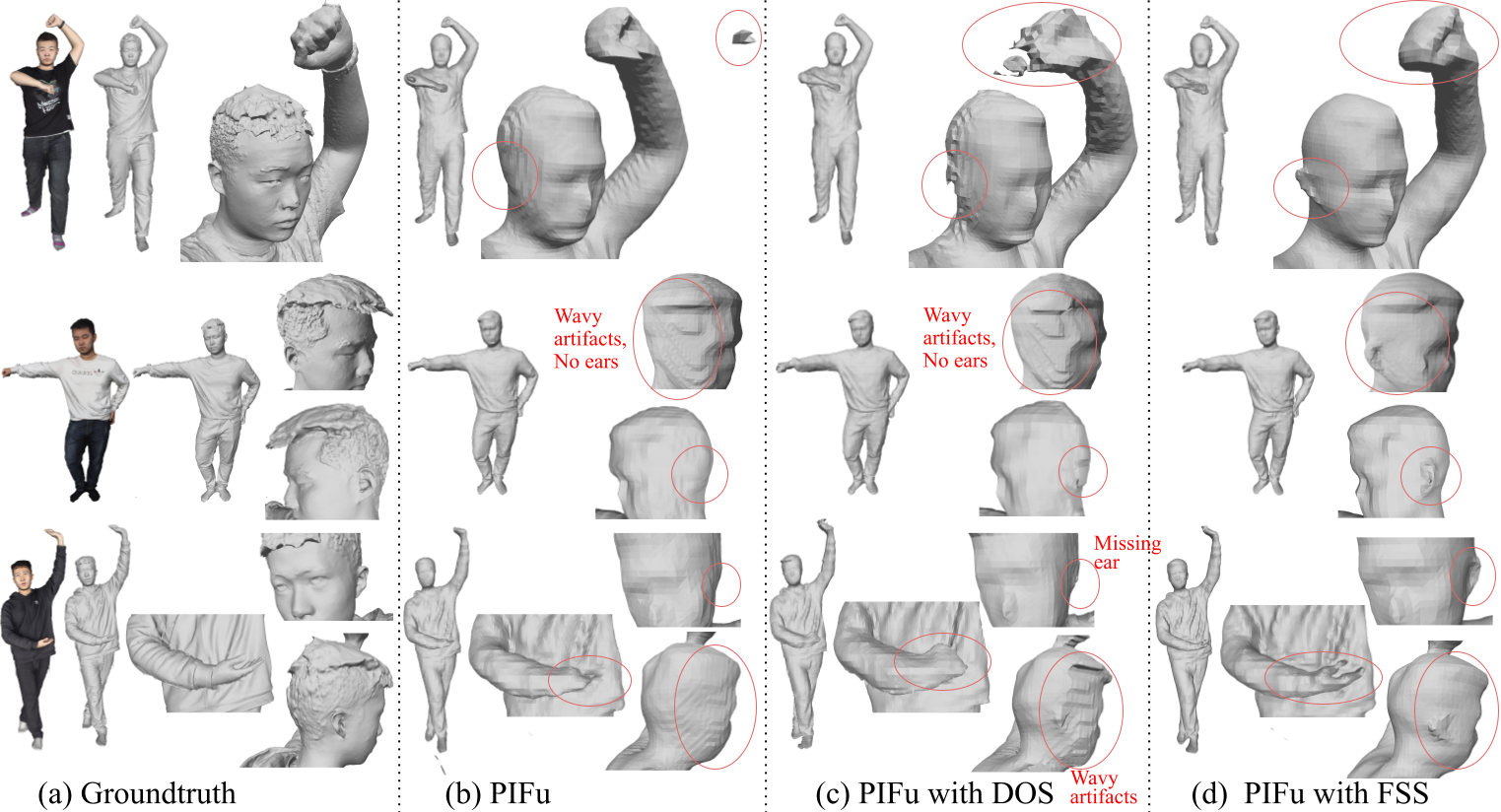}
\vspace{-6mm}
\caption{Evaluating our FSS scheme (w/o NSP w/o MTL)}
\label{fig:SmplxGuidedAblation}
\vspace{-4mm}
\end{figure}

We compare a PIFu trained with spatial sampling scheme, a PIFu trained with DOS scheme, and a PIFu trained with our FSS (w/o NSP w/o MTL). We present the qualitative results in Fig. \ref{fig:SmplxGuidedAblation}. We find that with our FSS (w/o NSP w/o MTL), a PIFu does not produce wavy, noisy artefacts and is much better at reconstructing thin features (e.g. ears, fingers). 

A quantitative evaluation can also be done using Tab. \ref{table:sotaQuantitative}. Comparing the rows of `PIFu', `PIFu + DOS', and `FSS w/o NSP w/o MTL (Ours)', we find that a PIFu trained with FSS is able to significantly outperforms both a PIFu trained with spatial sampling and a PIFu trained with DOS. 

Moreover, we did a further ablation study exploring the effects of including and excluding each of the five key features of FSS mentioned in Section \ref{ProximityguidedSamplingScheme_section}. Results can be seen quantitatively in Tab. \ref{table:FSS_FurtherAblation} and qualitatively in our Supp. Mat. As seen, each and every of the five key features is important.

\begin{figure}[t]

\centering
\begin{minipage}{.5\textwidth}
  \centering
  \includegraphics[width=0.98\linewidth]{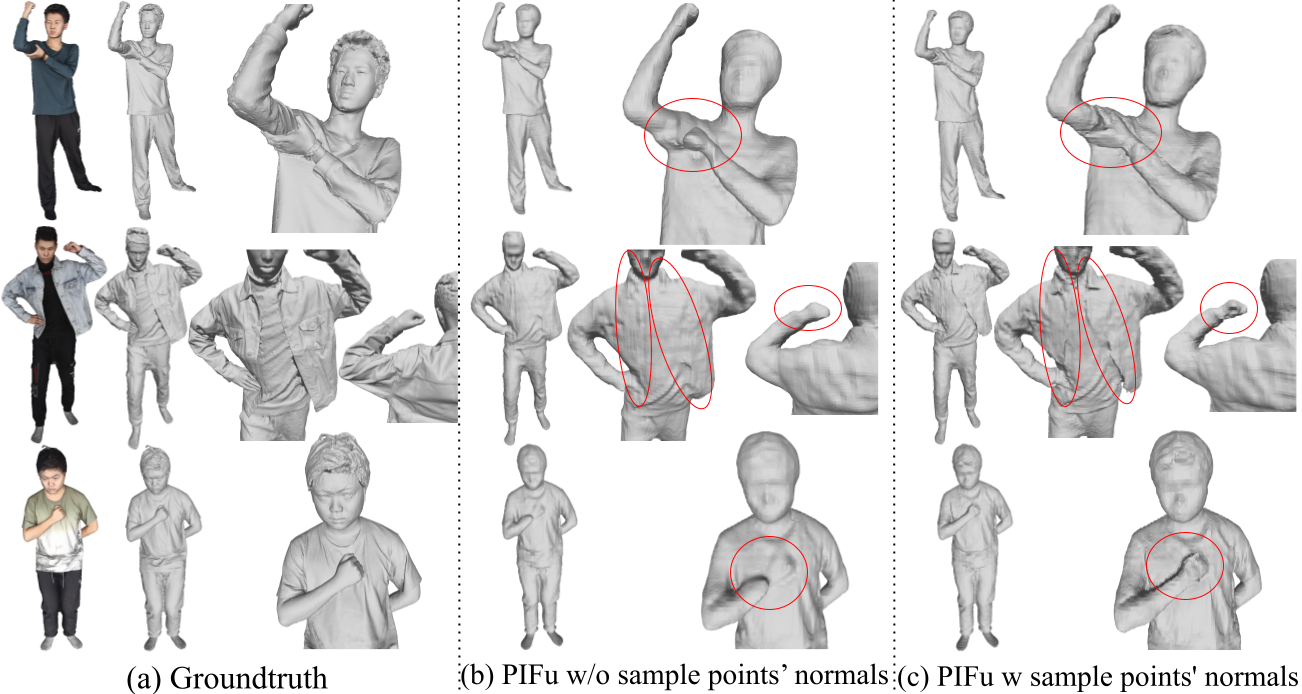} 
  \vspace{-2mm}
  \captionof{figure}{Effect of training with sample points' normals}
  \label{fig:PredNormAblation}
\end{minipage}%
\begin{minipage}{.5\textwidth}
  \centering
  \includegraphics[width=0.98\linewidth]{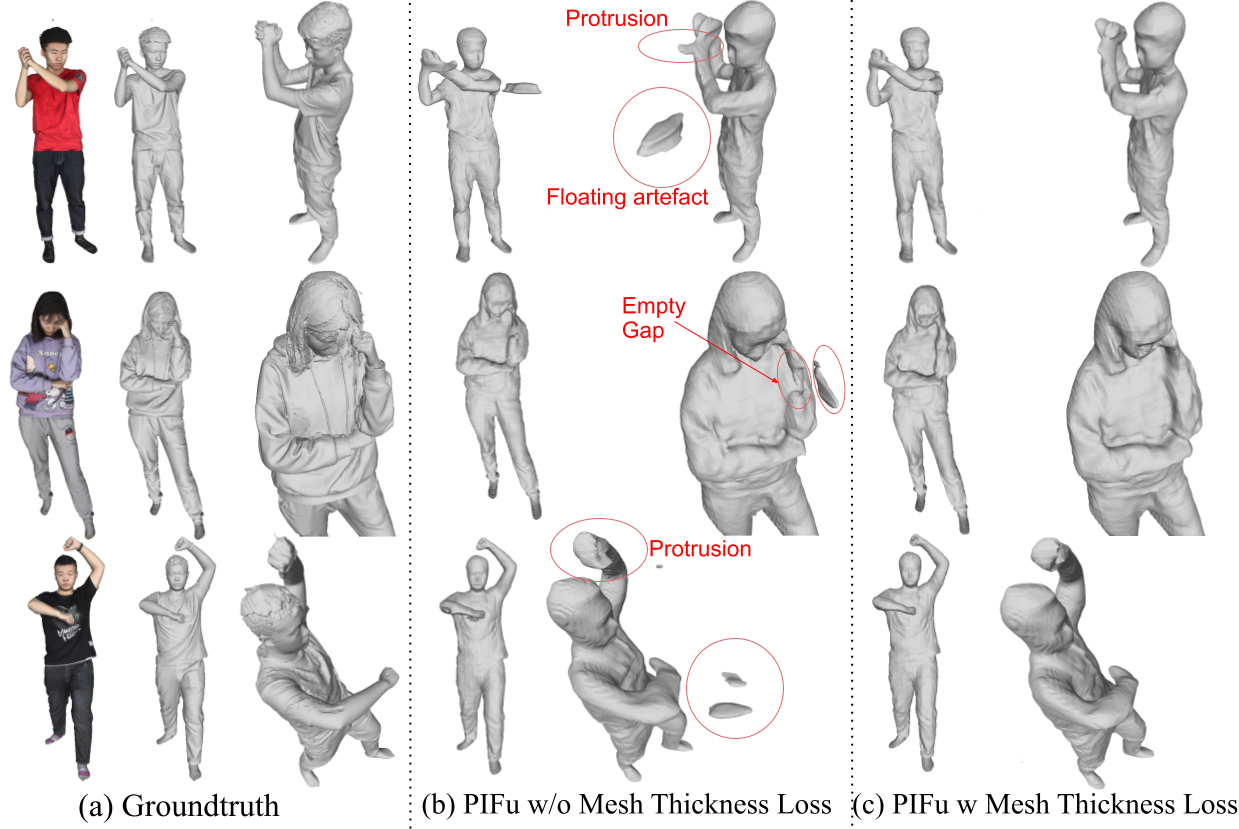} 
  \vspace{-2mm}
  \captionof{figure}{Evaluation of our mesh thickness loss signal}
  \label{fig:SDFFilterAblation}
\end{minipage}

\vspace{7mm}

\begin{minipage}{1.0\textwidth}
\centering
\includegraphics[width=0.95\textwidth]{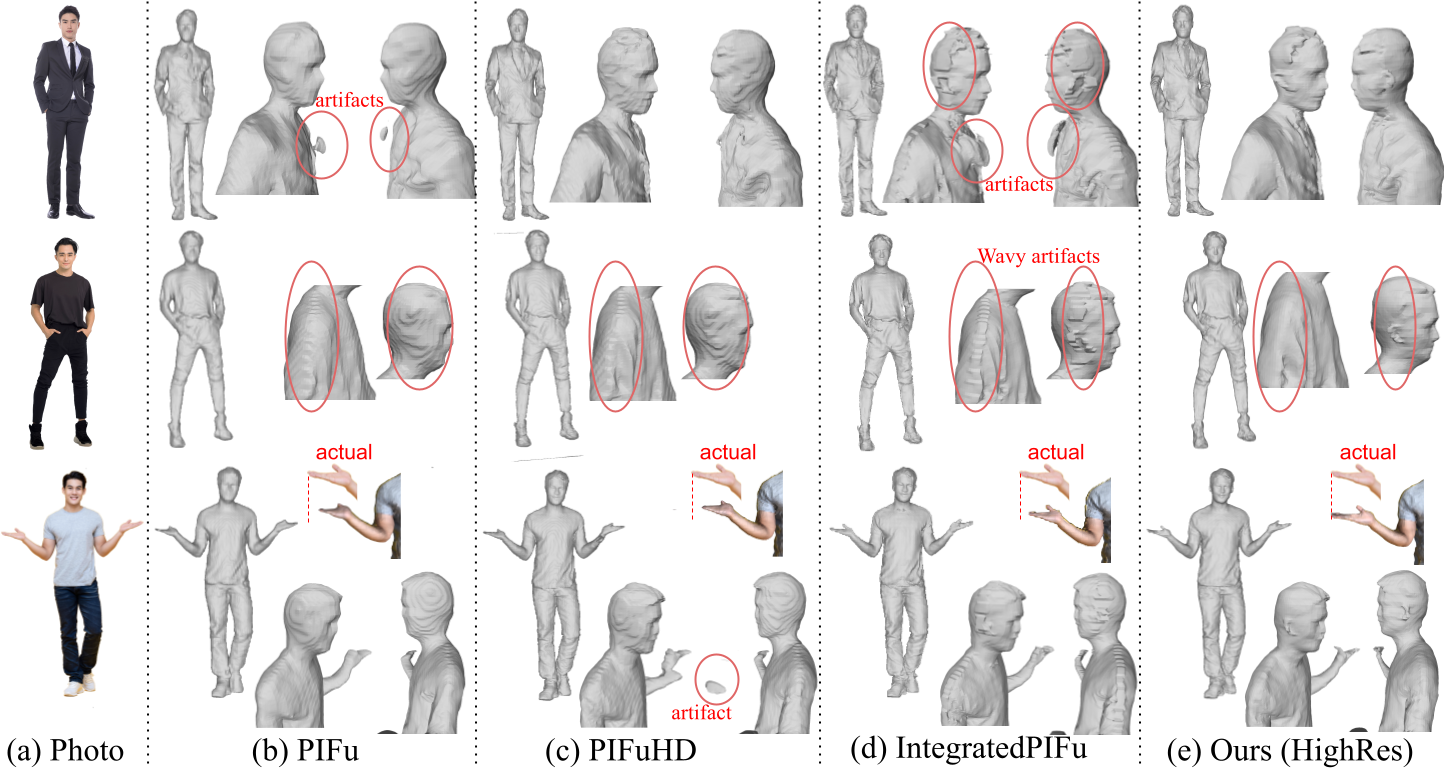}
\caption{Results with real images from Shutterstock}
\label{fig:InternetPhotos}

\end{minipage}

\end{figure}

\vspace{-1.1mm}
\paragraph{Evaluation of NSP}
To evaluate the usefulness of capitalizing on Normals of Sample Points (NSP) during the training process, we compare a PIFu trained without NSP with a PIFu that is trained with NSP. The results in Fig. \ref{fig:PredNormAblation} and Tab. \ref{table:normalAblation} show NSP improves both structure and details.

\vspace{-1.1mm}
\paragraph{Evaluation of MTL}
To evaluate the usefulness of our Mesh Thickness Loss (MTL), we compared a PIFu with a PIFu modified to use MTL. The results in Fig. \ref{fig:SDFFilterAblation} and Tab. \ref{table:SDFFilterAblation} show that MTL improves structural accuracy.

\section{Conclusion}

We have proposed Fine Structured-Aware Sampling (FSS), a novel sampling training scheme to train pixel-aligned implicit models for single-view human reconstruction. FSS allows models to capture and reconstruct thin but important human features effectively. FSS also shows how sample points' normals can be capitalized in the training process to further improve results.
Lastly, FSS also explained how a mesh thickness loss signal can be introduced with a slight reworking of the pixel-aligned implicit function framework.

\vspace{-2.0mm}
\begin{table}[h!]
\begin{center}
\caption{Evaluating FSS's five key features in THuman2.0}
\label{table:FSS_FurtherAblation}
\vspace{-3.5mm}
\resizebox{0.85\columnwidth}{!} {
\begin{tabular}{l|lll}
\hline 
Methods & CD (10\textsuperscript{-4}) & P2S (10\textsuperscript{-4}) & Normal (10\textsuperscript{-2}) \\
\hline
FSS & \textbf{5.004} & 3.965 & \textbf{8.052} \\
FSS w/o Twinned Query Points & 5.302 & \textbf{3.902} & 8.728 \\ 
FSS w/o Proximity-adapative displacement & 5.399 & 3.997 & 8.476 \\ 
FSS w/o Anchor query points & 5.279 & 4.021 & 8.326 \\
FSS w/o Counter query points & 5.187 & 4.502 & 8.854 \\
FSS w/o Smplx-guided Sampling & 5.135 & 4.133 & 8.761 \\
\hline
\end{tabular} 
}
\end{center}
\vspace{-6mm}
\end{table}

\vspace{-1.0mm}
\setlength{\tabcolsep}{4pt}
\begin{table}[h!]
\begin{center}
\caption{Ablation on using Normals of Sample Points (NSP)}
\vspace{-4mm}
\label{table:normalAblation}
\resizebox{1.0\columnwidth}{!} {
\begin{tabular}{l|lll|lll}
\hline 
 & \multicolumn{3}{c|}{THuman2.0} & \multicolumn{3}{c}{BUFF} \\
Methods & CD (10\textsuperscript{-4}) & P2S (10\textsuperscript{-4}) & Normal (10\textsuperscript{-2}) & CD (10\textsuperscript{3}) & P2S (10\textsuperscript{3}) & Normal (10\textsuperscript{-2}) \\
\hline
PIFu w/o NSP & 5.314 & 4.933 & 8.149 & 2.089 & 1.977 & 6.243 \\ 
\hline
PIFu w NSP & \textbf{4.991} & \textbf{4.041} & \textbf{7.905} & \textbf{2.041} & \textbf{1.841} & \textbf{6.100} \\ 
\hline
\end{tabular} 
}
\end{center}
\vspace{-6mm}
\end{table}
\setlength{\tabcolsep}{1.4pt}

\setlength{\tabcolsep}{4pt}
\begin{table}[h!]
\begin{center}
\caption{Ablation on using Mesh Thickness Loss (MTL)}
\vspace{-4mm}
\label{table:SDFFilterAblation}
\resizebox{1.0\columnwidth}{!} {
\begin{tabular}{l|lll|lll}
\hline 
 & \multicolumn{3}{c|}{THuman2.0} & \multicolumn{3}{c}{BUFF} \\
Methods & CD (10\textsuperscript{-4}) & P2S (10\textsuperscript{-4}) & Normal (10\textsuperscript{-2}) & CD (10\textsuperscript{3}) & P2S (10\textsuperscript{3}) & Normal (10\textsuperscript{-2}) \\
\hline
PIFu w/o MTL & 5.314 & 4.933 & 8.149 & 2.089 & 1.977 & 6.243 \\ 
\hline
PIFu w MTL & \textbf{5.131} & \textbf{3.927} & \textbf{8.009} & \textbf{1.953} & \textbf{1.739} & \textbf{6.036} \\ 
\hline
\end{tabular} 
}
\end{center}
\vspace{-8mm}
\end{table}
\setlength{\tabcolsep}{1.4pt}

\clearpage

\section*{Acknowledgements}
This research work is supported by the Agency for Science, Technology and Research (A*STAR) under its MTC Programmatic Funds (Grant No. M23L7b0021).

\bibliography{aaai24}

\end{document}